\documentclass{article}

\usepackage[preprint]{neurips_2023}

\usepackage[table,dvipsnames]{xcolor}
\usepackage{colortbl}
\usepackage{graphicx}
\usepackage{amsmath,amssymb} 
\usepackage{color}

\usepackage{natbib}
\PassOptionsToPackage{numbers, compress}{natbib}
\setcitestyle{square}
\setcitestyle{comma}

\usepackage{hyperref}
\hypersetup{
	colorlinks=true,
	citecolor=teal,
}
\usepackage{url}
\usepackage{booktabs}
\usepackage{microtype}      
\usepackage{multicol}
\usepackage{amsmath}
\usepackage{enumitem}
\usepackage{amssymb}
\numberwithin{equation}{section} 
\usepackage{wrapfig,lipsum,booktabs}
\usepackage{xcolor}
\usepackage{multirow}
\usepackage{float}
\usepackage{graphicx}
\usepackage{subcaption}
\usepackage{algpseudocode}  
\usepackage{algorithmicx,algorithm}
\usepackage{caption}
\usepackage{diagbox} 
\usepackage{colortbl}
\usepackage{xcolor}
\definecolor{rowcolor}{rgb}{0.9, 0.9, 0.9}
\usepackage{soul}

\graphicspath{ {images/} }

\usepackage{algorithmicx,algorithm}
\definecolor{commentcolor}{RGB}{110,154,155}   
\definecolor{functioncolor}{RGB}{200,2,127}   

\usepackage{pifont}
\usepackage{enumitem}
\let\oldding\ding
\renewcommand{\ding}[2][1]{\scalebox{#1}{\oldding{#2}}}
\newcommand{\ecar}[0]{\texttt{\small E-}\texttt{\normalsize CAR}}

\def\eg{\emph{e.g.}{}} 
\def\ie{\emph{i.e.}}

\title{\texttt{E-CAR}: Efficient Continuous Autoregressive Image Generation via Multistage Modeling}

\author{
\textbf{Zhihang Yuan$^{1,2}\thanks{Equal Contribution.}$~~~Yuzhang Shang$^{3 *}$~~~Hanling Zhang$^{1,2}$~~~Tongcheng Fang$^{1}$~~~Rui Xie$^{1,2}$} \\
\textbf{Bingxin Xu$^{3}$~~~Yan Yan$^{3}$~~~Shengen Yan$^{4}$~~~Guohao Dai$^{4,2}$~~~Yu Wang$^{1}$}\\
\\
\normalsize{$^{1}$Tsinghua University~~~~~$^{2}$Infinigence AI~~~~~$^{3}$Illinois Tech~~~~~$^{4}$Shanghai Jiao Tong University}
}

\begin{document}

\maketitle
\begin{abstract}
Recent advances in autoregressive (AR) models with continuous tokens for image generation show promising results by eliminating the need for discrete tokenization. However, these models face efficiency challenges due to their sequential token generation nature and reliance on computationally intensive diffusion-based sampling.
We present \ecar~(\textbf{E}fficient \textbf{C}ontinuous \textbf{A}uto-\textbf{r}egressive Image Generation via Multistage Modeling), an approach that addresses these limitations through two intertwined innovations: \raisebox{-1.1pt}{\ding[1.1]{182\relax}} a stage-wise continuous token generation strategy that reduces computational complexity and provides progressively refined token maps as hierarchical conditions, and \raisebox{-1.1pt}{\ding[1.1]{183\relax}} a multistage flow-based distribution modeling method that transforms  only partial-denoised distributions at each stage comparing to complete denoising in normal diffusion models.
Holistically, \ecar~operates by generating tokens at increasing resolutions while simultaneously denoising the image at each stage. This design not only reduces token-to-image transformation cost by a factor of the stage number but also enables parallel processing at the token level.
Our approach enhances computational efficiency and aligns naturally with image generation principles by operating in continuous token space and following a hierarchical generation process from coarse to fine details.
Experimental results demonstrate that \ecar~achieves comparable image quality to DiT~\citep{peebles2023scalable} while requiring 10$\times$ FLOPs reduction and 5$\times$ speedup to generate a 256$\times$256 image.
\end{abstract}

\section{Introduction}
\label{sec:intro}

Autoregressive (AR) Large Language Models (LLMs)~\citep{vaswani2017attention,achiam2023gpt,touvron2023llama,touvron2023llama2,jiang2023mistral} have demonstrated remarkable capabilities in natural language generation, driven by scaling laws~\citep{kaplan2020scaling,hoffmann2022training}. 
Inspired by these advancements, the computer vision community has been striving to develop large autoregressive and world models~\citep{zhou2024transfusion,team2024chameleon,liu2024lumina} for visual generation. This effort is initiated by VQ-GAN~\citep{esser2021taming} and further advanced by its successors~\citep{lee2022autoregressive,sun2024autoregressive,tian2024visual,team2024chameleon}, showcasing the potential of AR models in visual generation.

Traditional AR image generation models employ a visual tokenizer to discretize continuous images into grids of 2D tokens, which are then flattened into a 1D sequence for AR learning (see Fig.~\ref{fig:intro}.b). However, recent research~\citep{li2024autoregressive,tschannen2023givt,zhou2024transfusion,fan2024fluid} demonstrates that discrete tokenization is not only unnecessary but potentially detrimental to AR models' generation capabilities. These works show that AR image generation can be formulated continuously, modeling per-token probability distributions on continuous-valued domains. For instance, several approaches~\citep{li2024autoregressive,zhou2024transfusion,fan2024fluid} leverage diffusion models for representing arbitrary probability distributions, where the model autoregressively predicts a continuous vector for each token as conditioning for a denoising network (see Fig.~\ref{fig:intro}.c). We refer to this paradigm as \textit{continuous AR} models.

Despite their advantages, continuous AR models, such as MAR~\citep{li2024autoregressive}, face significant efficiency challenges. Generating a 256$\times$256 image with MAR using one NVIDIA A5000 GPU requires more than 30 seconds, making these models impractical for real-time applications. We identify two main efficiency bottlenecks:
\raisebox{-1.1pt}{\ding[1.1]{182\relax}} Token-by-token generation: Similar to text generation in LLMs, the sequential token-by-token nature of generation limits efficiency~\citep{yuan2024llm}.
\raisebox{-1.1pt}{\ding[1.1]{183\relax}} Diffusion-based sampling: The sampling process inherits the inefficiencies of diffusion models~\citep{song2020denoising,song2020score}, requiring hundreds of denoising network inferences for token-to-image sampling.
Beyond the direct computational costs, these issues significantly constrain parallelization potential, as the token generation process must proceed sequentially.

\begin{wrapfigure}{r}{0.45\textwidth}
\vspace{-0.2in}
    \centering
    \includegraphics[width=0.95\linewidth]{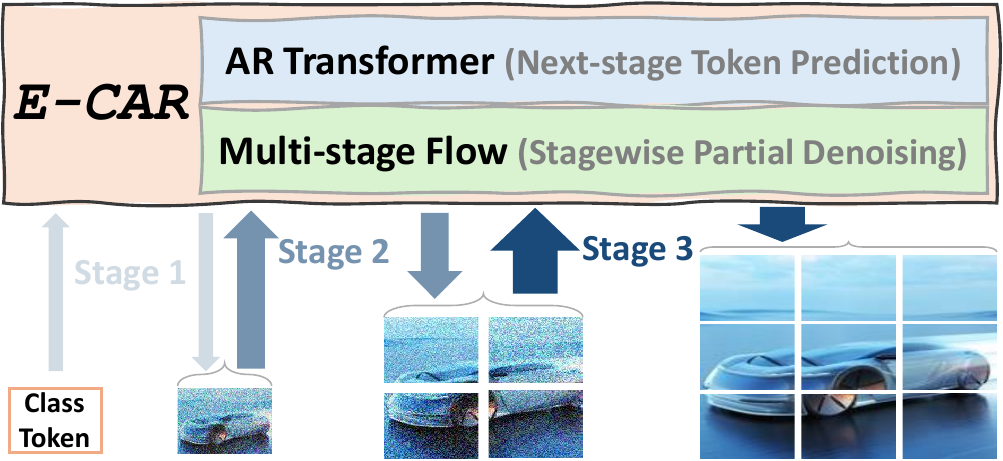}
    \caption{\textbf{High-level Idea of \ecar.} The model progressively generates tokens at increasing resolutions, while correspondingly denoising the image at each stage. By combining stage-by-stage continuous token generation with multistage flow-based image synthesis, \ecar~achieves efficient continuous autoregressive image generation while maintaining high visual quality.}
    \vspace{-0.25in}
    \label{fig:brief_idea}
\end{wrapfigure}

In this work, we propose Efficient Continuous Autoregressive Image Generation via Multistage Modeling (\ecar) to address these limitations while maintaining generative capability. As shown in Fig.~\ref{fig:brief_idea}, \ecar~progressively generate tokens at increasing resolutions while simultaneously transport towards the image distribution via a multistage flow at each stage. 
Our approach features two intervened innovations targeting the efficiency bottleneck:
\raisebox{-1.1pt}{\ding[1.1]{182\relax}} \textbf{Stage-wise Progressive Token Map Generation:} A hierarchical AR transformer architecture that generates continuous tokens at stage-wisely increasing resolutions (\ie, continuous token map), reducing token generation computation by enabling parallel processing within each stage.
\raisebox{-1.1pt}{\ding[1.1]{183\relax}} \textbf{Multistage Flow-based Distribution Modeling:} A flow-based method that transforms continuous token distributions at multiple stages, requiring only partial transport at each resolution level instead of complete denoising processes. This design reduces continuous detokenization computation proportionally to the number of stages.
The stage-wise design of our approach enables parallel token sampling, as tokens are generated in resolution-specific stacks. From a flow matching perspective~\citep{ma2024sit,lipman2022flow,liu2022flow,liu2022rectified}, our multistage approach can be viewed as using the AR model to guide the flow model's visual generation process, creating a complementary relationship between continuous token generation and flow-based image synthesis.

Importantly, \ecar's multistage continuous AR solution aligns naturally with image generation's pyramidal principles~\citep{pernias2023wurstchen,saharia2022photorealistic}. By operating in a continuous token space, it better reflects the inherent structure of images—a widely accepted inductive bias in computer vision~\citep{li2024autoregressive}. The increasing resolution generation mirrors the hierarchical nature of visual information, progressing from coarse structures to fine details, enabling effective multi-scale image generation while maintaining high quality~\citep{jin2024pyramidal}.
To the best of our knowledge, this is the first work that integrates continuous AR and flow-based method in a multistage manner, offering a novel approach to efficient, high-quality visual generation.

\section{Related Work}
\label{sec:related}
\noindent\textbf{Autoregressive Visual Generation} models initially operated on sequences of pixels~\citep{gregor2014deep,van2016pixel,chen2018pixelsnail,chen2020generative}. While early implementations primarily utilized RNNs and CNNs as base models, Transformers~\citep{vaswani2017attention} have recently become the dominant architecture for these tasks~\citep{esser2021taming}.
VQGAN~\citep{esser2021taming} pioneered the use of Transformers for autoregressive image generation. Specifically, it employs a GPT-2 decoder-only Transformer~\citep{radford2018improving} to generate tokens in a raster-scan order, similar to how ViT~\citep{dosovitskiy2020image} serializes 2D images into 1D patches. Building upon this paradigm, \citep{razavi2019generating,lee2022autoregressive} extend the approach by incorporating multiple scales or stacked codes. More recently, LlamaGen~\citep{sun2024autoregressive}, based on the popular open-source LLM architecture Llama~\citep{touvron2023llama,touvron2023llama2}, scales the Transformer to 3B parameters and demonstrated impressive results in text-to-image synthesis.
In parallel, recent advancements have focused on continuous-valued tokens in AR models. GIVT~\citep{tschannen2023givt} represents token distributions using Gaussian mixture models with a pre-defined number of mixtures. MAR~\citep{li2024autoregressive} leverages the effectiveness of the diffusion process for modeling arbitrary distributions. We categorize these AR methods that do not use discrete tokenizers as \textit{continuous AR} methods.

Our work focuses on improving the efficiency of continuous AR models. To achieve this, we propose using a multistage AR Transformer to generate continuous tokens stage-by-stage, coupled with a multistage flow-based method to model the token distributions.

\noindent\textbf{Diffusion Models and Flow Matching Models.}
Diffusion models~\citep{sohl2015deep,ho2020denoising,song2020score} have demonstrated remarkable success across various generative modeling tasks, such as the generation of images, videos, and audio~\citep{croitoru2023diffusion}. Predominantly, these models utilize stochastic differential equations (SDEs) to model the diffusion and denoising processes. \citep{song2020score,song2020denoising} also converted these SDEs into ordinary differential equations (ODEs) that preserve the marginal probability distributions, thereby accelerating the denoising process. Recent research~\citep{liu2022flow,liu2022rectified,lipman2022flow} has introduced novel methods for directly learning probability flow ODEs, employing both linear and nonlinear interpolation techniques between distributions. These ODE-based approaches rival the performance of conventional diffusion models while substantially reducing the number of inference steps required.
The practical applications of flow matching have expanded beyond theoretical frameworks, with recent investigations applying these methods to larger datasets and more complex tasks. For example, \citep{liu2023instaflow} showcased the Rectified Flow pipeline's capability to produce high-fidelity, one-step generation in large-scale text-to-image (T2I) diffusion models. This advancement has paved the way for extremely rapid T2I foundation models trained exclusively through supervised learning. The influence of these developments is evident in cutting-edge T2I systems, such as Stable Diffusion 3~\citep{esser2024scaling}, which incorporates reflow as a fundamental component of its generation process.

Our work focuses on continuous AR, but we formulate the distribution estimation under the framework of multistage flow matching, offering a more efficient method for continuous distribution matching.

\section{\ecar: Efficient Continuous Auto-regressive Image Generation via Multistage Modeling}
\label{sec:method}
We first review the basic pipeline of continuous autoregressive (AR) visual generation models (Sec.\ref{subsec:prelim}). 
Next, we introduce our \textbf{E}fficient \textbf{C}ontinuous \textbf{AR} (\textbf{\ecar}) image generation model.
The development and optimization of \ecar~include three key advancements: 
(1) a stage-by-stage continuous token generation AR module (Sec.\ref{subsec:h-ar}); (2) an multistage flow model for fast continuous token recovery (Sec.~\ref{subsec:flow}); and (3) the multistage loss that enhances training stability (Sec.\ref{subsec:loss}). 
Finally, we provide a straightforward explanation of why \ecar~can be both efficient and effective (Sec.\ref{subsec:discuss}).

\subsection{Preliminaries: Continuous Autoregressive Visual Generation}
\label{subsec:prelim}
The review of continuous AR visual generation models focuses on their main components and training procedures. 
In this review, we highlight the inefficiencies associated with existing continuous AR models.

\noindent\textbf{Image Tokenization and Detokenization.} 
Images are inherently 2D continuous signals. Pioneering AR studies~\citep{esser2021taming,sun2024autoregressive,tian2024visual,yu2023magvit,yu2023language} apply AR modeling to images via next-token prediction, mimicking AR in language models~\citep{vaswani2017attention}. This approach requires two key steps: (1) tokenizing an image into discrete tokens, and (2) modeling the tokens' unidirectional generation process. 

For tokenization, a quantized autoencoder such as VQGAN~\citep{esser2021taming} is often used to convert the image feature map $\mathbf{f} \in \mathbb{R}^{(h \cdot w) \times C}$ to discrete tokens $\mathbf{z} \in [V]^{(h \cdot w)}$:
\begin{equation}
\mathbf{f} = \mathcal{E}(\mathbf{x}), \quad \mathbf{z} = Q(\mathbf{f}),
\end{equation}
where $\mathbf{x} \in \mathbb{R}^{h^{\prime} \times w^{\prime} \times 3}$ denotes the raw image, $\mathcal{E}(\cdot)$ is an encoder, and $Q(\cdot)$ is a quantizer. The quantizer typically includes a learnable $\text{Codebook} \in \mathbb{R}^{V \times C}$ containing $V$ vectors with $C$ dimensions. The quantization process $\mathbf{z} = Q(\mathbf{f})$ maps each feature vector $\mathbf{f}^{i}$ to the code index $\mathbf{z}^{i}$ of its nearest code w.r.t. the Euclidean distance:
\begin{equation}
\mathbf{z}_{j} = \arg\min_{v\in[V]} \|\text{lookup}(\text{Codebook}, v) - \mathbf{f}_{j}\|_2,
\label{eq:vq_lookup}
\end{equation}
where lookup$(\text{Codebook}, v)$ means taking the $v$-th vector in codebook, $\mathbf{z}_{j}$ and $\mathbf{f}_{j}$ means $j$-th discrete token and feature map, respectively. 

Quantized autoencoder training involves reconstructing the image and minimizing a compound loss:
\begin{equation}
    \begin{aligned}
        \label{eq:ar_generation}
        &\hat{\mathbf{f}} = \text{lookup}(\mathbf{Z}, \mathbf{z}), \quad \hat{\mathbf{x}} = \mathcal{D}(\hat{\mathbf{f}}), \\
        \mathcal{L} = \|\mathbf{x} - &\hat{\mathbf{x}}\|_2 + \|\mathbf{f} - \hat{\mathbf{f}}\|_2 + \lambda_P \mathcal{L}_P(\hat{\mathbf{x}}) + \lambda_G \mathcal{L}_G(\hat{\mathbf{x}}),
    \end{aligned}
\end{equation}
where $\mathcal{D}(\cdot)$ is the decoder, $\mathcal{L}_P(\cdot)$ is a perceptual loss, and $\mathcal{L}_G(\cdot)$ is a discriminative loss. Using this tokenizer, a continuous image $\mathbf{x}$ can be discretized into a sequence of categorical tokens $\mathbf{z} = (z_1, \ldots, z_I)$. Conversely, generated tokens can be detokenized and then decoded back into a continuous image (i.e., realizing image generation).

\begin{figure*}[t]
    \centering
    \includegraphics[width=0.98\linewidth]{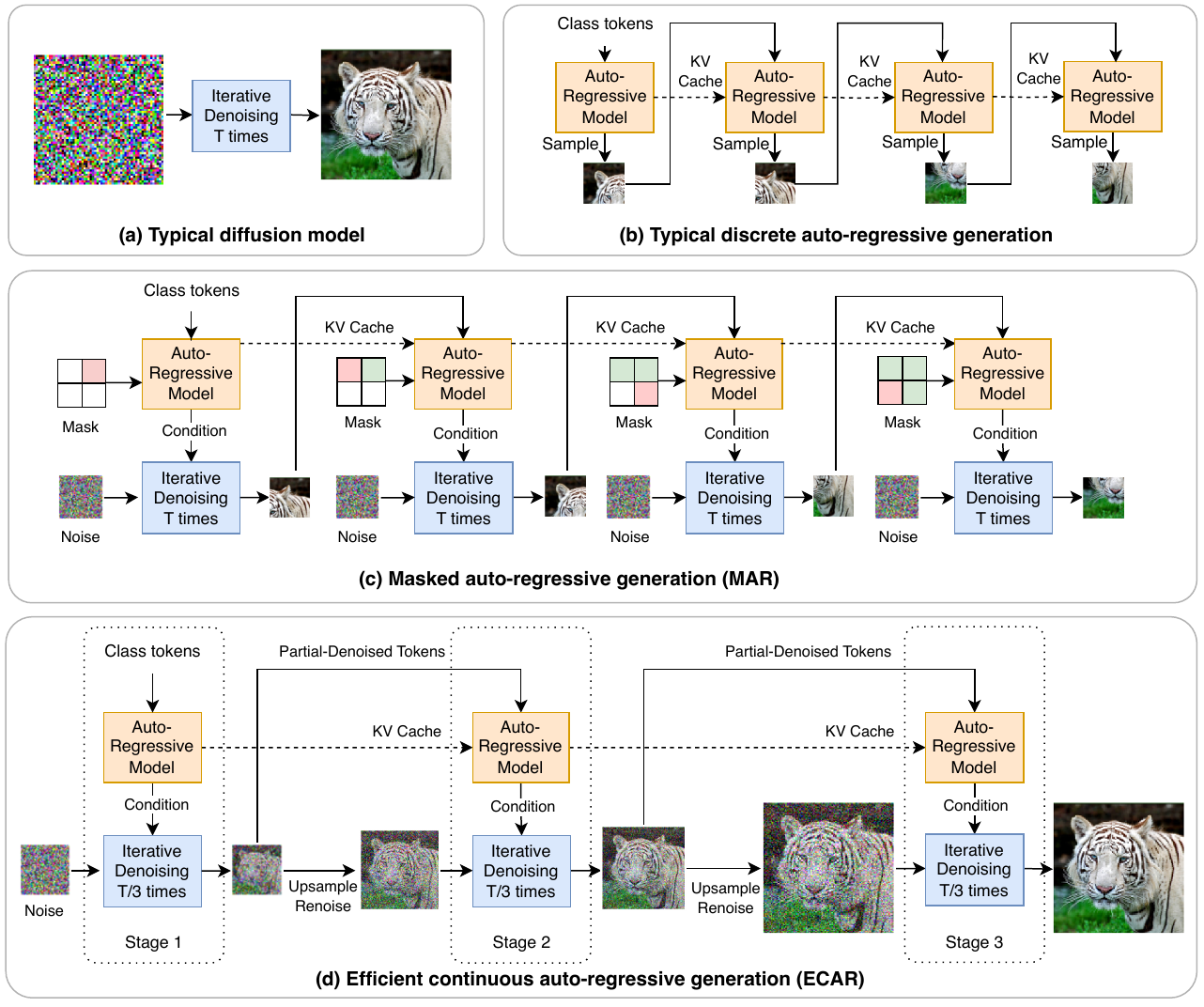}
    \caption{
    \textbf{(a)} Diffusion/Flow-matching model: Generates images through multiple iterations of denoising/velocity network inference. \textbf{(b)} Traditional AR Transformer: Sequentially generates discrete tokens, followed by codebook-based detokenization. \textbf{(c)} Continuous Masked AR~\citep{li2024autoregressive}: Sequentially produces continuous tokens, which are transformed into image patches via a diffusion model. \textbf{(d)} \textbf{\ecar}: Introduces two key innovations: 
    multi-stage continuous token generation and \textbf{(Sec.\ref{subsec:flow}) multi-stage flow}
     for efficient continuous token generation and token-to-image detokenization, respectively. Using the upsample and renoise technique~\citep{jin2024pyramidal}, we can correspondingly reduce the number of steps for flow matching at each stage, enhancing the efficiency of the continuous token detokenization process.}
    \label{fig:intro}
    \vspace{-0.1in}
\end{figure*}

\noindent\textbf{Autoregressive Token Generation} model predicts tokens sequentially, by factorizing the likelihood of a sequence $\mathbf{z} = (z_1, \ldots, z_I)$ as:
\begin{equation}
p(z_1, z_2, \ldots, z_I) = \prod_{t=1}^T p(z_i | z_1, z_2, \ldots, z_{i-1}).
\label{eq:autoregressive}
\end{equation}
In the traditional AR image generation models~\citep{esser2021taming,sun2024autoregressive,tian2024visual,yu2023magvit,yu2023language} (see Fig.\ref{fig:intro}), $\{z_i\}_{t=1}^T$ are discrete tokens. 
However, the discrete tokens $\{z_i\}_{t=1}^T$ are not directly generated by the AR model. In fact, the autoregressive model produces a continuous-valued $C$-dim vector $\mathbf{z}_i \in \mathbb{R}^C$, which is then projected by a $V$-way classifier matrix $\mathbf{W} \in \mathbb{R}^{C\times V}$ to $z_i \in [V]$.
In other words, one can use Eq.\ref{eq:autoregressive} and an additional projector to generate a sequence of categorical tokens, and then use Eq.\ref{eq:ar_generation} to decoder back to the image space. 

\noindent\textbf{Continuous AR.}
Recent work \citep{li2024autoregressive,tschannen2023givt} formulate the probability distribution $p(\mathbf{f}_i|\mathbf{z}_i)$ to model the detokenization for continuous tokens, where $\mathbf{f}_i \in \mathbb{R}^C$ is the $t$-th image patch's feature map and $\mathbf{z}_i \in \mathbb{R}^D$ is the corresponding AR model's output \textit{continuous token}. 
MAR~\citep{li2024autoregressive} uses diffusion models to define the loss function and sampler, as shown in Fig.\ref{fig:intro} (c).
MAR and its follow-up work~\citep{fan2024fluid} achieve state-of-the-art performance in image generation, showing the promising potential of continuous AR in the visual generation.

\noindent\textbf{Efficiency Bottlenecks of Continuous AR Models.}
Continuous AR models face two main efficiency challenges:
\raisebox{-1.1pt}{\ding[1.1]{182\relax}} Token-by-token generation: Similar to LLMs, the sequential nature of generation limits efficiency~\citep{yuan2024llm}. In these models, each token is produced based on all the previously generated tokens, which means that the process cannot be easily parallelized. This sequential dependency results in longer computation times, especially for generating lengthy tokens (corresponding to high-resolution images or videos).
\raisebox{-1.1pt}{\ding[1.1]{183\relax}} Diffusion sampling: \citep{song2020denoising} diffusion models require a large number of iterative denoising steps to generate a single sample, each of which must be executed sequentially \citep{song2020denoising}. 

We propose a hierarchical framework for addressing these issues, in which continuous token generation and continuous detokenization are realized in a hierarchical manner. By doing so, we are able to improve the efficiency of the continuous AR model as well as maintain its effectiveness.

\subsection{Multistage Autoregressive Token Generation}
\label{subsec:h-ar}
Targeting the \raisebox{-1.1pt}{\ding[1.1]{182\relax}} efficiency challenge, we introduce a stage-by-stage continuous tokens AR generation model to enhance the efficiency of token generation. Our method reimagines autoregressive image modeling by transitioning from a ``next-token prediction'' paradigm to a ``next-stage prediction'' strategy, as illustrated in Figs.\ref{fig:brief_idea} and \ref{fig:intro}. 
In this framework, the autoregressive unit evolves from a single token to an \textit{entire token map}. The process begins by encoding a feature map $\mathbf{M} = [\mathbf{z}_i, \cdots, \mathbf{z}_{(h \cdot w)}] \in \mathbb{R}^{(h \cdot w) \times C}$ into a sequence of $S$-stage token maps $(\mathbf{M}_1, \mathbf{M}_2, \ldots, \mathbf{M}_S)$. Each successive map increases in resolution from $h_1 \times w_1$ to $h_S \times w_S$, with the final map $m_S$ matching the original feature map's dimensions $(h \cdot w)$. The autoregressive likelihood can be formulated as:
\begin{equation}
    p(\mathbf{M}_1, \mathbf{M}_2, \ldots, \mathbf{M}_S) = \prod_{s=1}^S p(\mathbf{M}_s \mid \mathbf{M}_1, \mathbf{M}_2, \ldots, \mathbf{M}_{s-1}).
\end{equation}
Here, each autoregressive unit $\mathbf{M}_s \in \mathbb{R}^{(h_s \cdot w_s) \times C}$ represents the token map at stage $s$, encompassing $h_s \cdot w_s$ continuous tokens. The preceding sequence $(\mathbf{M}_1, \mathbf{M}_2, \ldots, \mathbf{M}_{s-1})$ functions as a ``prefix''~\citep{yuan2024llm} for $m_s$. 
During the $s$-th autoregressive step, our model generates distributions for all $h_s \cdot w_s$ tokens in $m_s$ concurrently. This parallel generation is conditioned on both the prefix and the stage-specific positional embedding map. We term this ``next-stage prediction'' approach Multi-stage Continuous Autoregressive, as depicted in Fig.~\ref{fig:intro}.

\noindent\textbf{Efficiency Benefit.} 
The designed multi-stage continuous token generation offers significant computational advantages over the vanilla continuous AR. Theoretically, our method achieves a time complexity of $\mathcal{O}(n^2)$, where $n$ is the number of tokens representing an image. In contrast, the vanilla continuous AR method has a time complexity of $\mathcal{O}(n^3)$. This improvement results in our approach being $n$ times faster than the vanilla method. More details can be found in the Supplementary Materials. 
We would like to highlight that this efficiency gain becomes increasingly significant as we move towards long-sequence token generation (\eg, higher-resolution images or video generation).

\subsection{Multistage Flow Matching for Fast Sampling from Continuous Token}
\label{subsec:flow}

After we get the set of continuous tokens $\{\mathbf{M}_s\}_{s=1}^{S}$, we hope to get a sampler that can draw samples from the distribution $\mathbf{F}_S \sim p(\mathbf{F}_S|\mathbf{M}_S)$ at inference time, \ie, generate image patches' embeddings based on the output tokens. 
Targeting the \raisebox{-1.1pt}{\ding[1.1]{183\relax}} efficiency challenge, we model this sampler under the framework using multistage flow~\citep{liu2022flow,lipman2022flow,liu2023instaflow,xie2024towards} to improve the continuous detokenization efficiency. 

\noindent\textbf{Problem Formulation.}
Consider the continuous-valued token map at $S$-th stage $\mathbf{M}_S \in \mathbb{R}^{(h_S \cdot w_S) \times D}$, and the ground-truth image patch embeddings $\mathbf{F}^{(S)} \in \mathbb{R}^{(h_S \cdot w_S) \times C}$ to be predicted at the this stage. Our goal is to model a probability distribution of $\mathbf{F}_S$ conditioned on $\mathbf{M}_S$, that is, $p(\mathbf{F}^{(S)}|\mathbf{M}_S)$. In the context of optimal transport, we hope to find a velocity model, optimal transport from $\mathbf{n}|\mathbf{M}_S \sim \pi_0$ to $\mathbf{F}^{(S)}|\mathbf{M}_S \sim \pi_1$. For simple sampling, $\pi_0$ commonly be set as a normal distribution.

\noindent\textbf{Flow Matching}~\citep{liu2022rectified,liu2022flow,ma2024sit,jin2024pyramidal} is a unified ODE-based framework for generative modeling and domain transfer. It provides an approach for learning a transport mapping $T$ between two distributions $\pi_0$ and $\pi_1$. 
Specifically, flow Matching learns to transfer $\pi_0$ to $\pi_1$ via an ordinary differential equation (ODE), or flow model
\begin{equation}
    \frac{d\mathbf{F}^{(S)}_t}{dt} = v_{\theta}((\mathbf{F}^{(S)}), t\mid \text{cond}=\mathbf{M}_S), \quad \text{initialized from }  \mathbf{F}^{(S)}_0 \sim \pi_0=\mathcal{N}(\mathbf{0},\mathbf{I}), \text{s.t.} \mathbf{F}^{(S)}_1 \sim \pi_1,
\label{eq:flow_ode}
\end{equation}
where $\text{cond}$ is the condition variable, it can be a class or a text embedding, and $v_{\theta}: \mathbb{R}^d \times [0,1] \times \mathbb{R}^D \to \mathbb{R}^d$ is a velocity field, learned by minimizing a mean square objective:
\begin{equation}
        \min_{v_{\theta}} \mathbb{E}_{((F_S)_0, (F_S)_1) \sim \gamma, z \sim \mathcal{D}_{\text{cond}}} \left[ \int_0^1  \left\|  \frac{d}{dt}\mathbf{F}^{(S)}_t - v_{\theta}(\mathbf{F}^{(S)}_t, t\mid \mathbf{M}_S) \right\|^2 dt \right],
\label{eq:rectflow_loss}
\end{equation}
with $\mathbf{F}^{(S)}_t = \phi(\mathbf{F}^{(S)}_0, \mathbf{F}^{(S)}_1, t)$where $\mathcal{D}_{\text{cond}}$ is the collection of conditions, $\mathbf{F}^{(S)}_t = \phi(\mathbf{F}^{(S)}_0, \mathbf{F}^{(S)}_1, t)$ is any time-differentiable interpolation between $\mathbf{F}^{(S)}_0$ and $\mathbf{F}^{(S)}_1$, with $\frac{\mathrm{d}}{\mathrm{d}t}\mathbf{F}^{(S)}_t = \partial_t\phi(\mathbf{F}^{(S)}_0, \mathbf{F}^{(S)}_1, t)$. The $\gamma$ is any coupling of $(\pi_0, \pi_1)$. A simple example of $\gamma$ is the independent coupling $\gamma = \pi_0 \times \pi_1$, which can be sampled empirically from unpaired observed data from $\pi_0$ and $\pi_1$. Usually, $v_{\theta}$ is parameterized as a deep neural network and Eq.\ref{eq:rectflow_loss} is solved approximately with stochastic gradient methods. Different specific choices of the interpolation process $\mathbf{F}^{(S)}_1$ result in different algorithms. As shown in \citep{liu2022rectified}, the commonly used denoising diffusion implicit model (DDIM)~\citep{song2020denoising} and the probability flow ODEs of \citep{song2020score} correspond to $\mathbf{F}^{(S)}_t = \alpha_t \mathbf{F}^{(S)}_0 + \beta_t \mathbf{F}^{(S)}_1$, with specific choices of time-differentiable sequences $\alpha_t, \beta_t$ (see \citep{liu2022rectified,liu2023instaflow} for details). In rectified flow~\citep{liu2022rectified}, the authors suggested a simpler choice of
\begin{equation}
\begin{aligned}
        \mathbf{F}^{(S)}_t = (1-t)\mathbf{F}^{(S)}_0 + t\mathbf{F}^{(S)}_1 
    \implies \frac{\mathrm{d}}{\mathrm{d}t}\mathbf{F}^{(S)}_t = \mathbf{F}^{(S)}_1 - \mathbf{F}^{(S)}_0.
\end{aligned}
\end{equation}

\noindent\textbf{Simple Flow Sampler.}
Once we obtain a well-trained velocity model $v_{\theta}$ through optimization of Eq.~\ref{eq:rectflow_loss}, we can sample image patches's embedding $\mathbf{F}_S)$ conditioned on the continuous token map $\mathbf{M}_S)$. The sampling process involves approximating the ODE in Eq.~\ref{eq:flow_ode} using numerical methods.
The most common approach for this approximation is the forward Euler method, which discretizes the continuous ODE into finite steps:
\begin{equation}
\begin{aligned}
        \mathbf{F}^{(S)}_{t+\Delta t} = \mathbf{F}^{(S)}_t + \Delta t \cdot v_{\theta}(\mathbf{F}^{(S)}_t, t \mid \text{cond}=\mathbf{M}_S),
\end{aligned}
\label{eq:flow_sampling}
\end{equation}
where $t \in \{0, \frac{1}{N}, \frac{2}{N}, \ldots, \frac{N-1}{N}\}$ and $\Delta t = \frac{1}{N}$ is the step size, $N$ is the total number of simulation steps, $\mathbf{F}^{(S)}_t$ represents the state at time $t$, and $v(\mathbf{F}^{(S)}_t, t)$ is the velocity predicted by our trained model. 
The sampling process starts at $t=0$ and iteratively applies this update equation $N$ times until reaching $t=1$, thereby transforming the initial noise distribution into the desired sample distribution. 

However, directly using full-size token map $\mathbf{M}_S$ to generate full-size feature map $\mathbf{F}^{(S)}$ in the final stage is not optimal for efficiency, specifically in our stage-wise token generation case. We hope to fully leverage the low-resolution token maps.
Therefore, to leverage our multi-stage token generation framework and accelerate sampling, we propose a multi-stage sampling strategy that aligns with the hierarchical nature of our token generation process. 

\begin{figure*}[tb]
    \centering
    \includegraphics[width=1.05\linewidth]{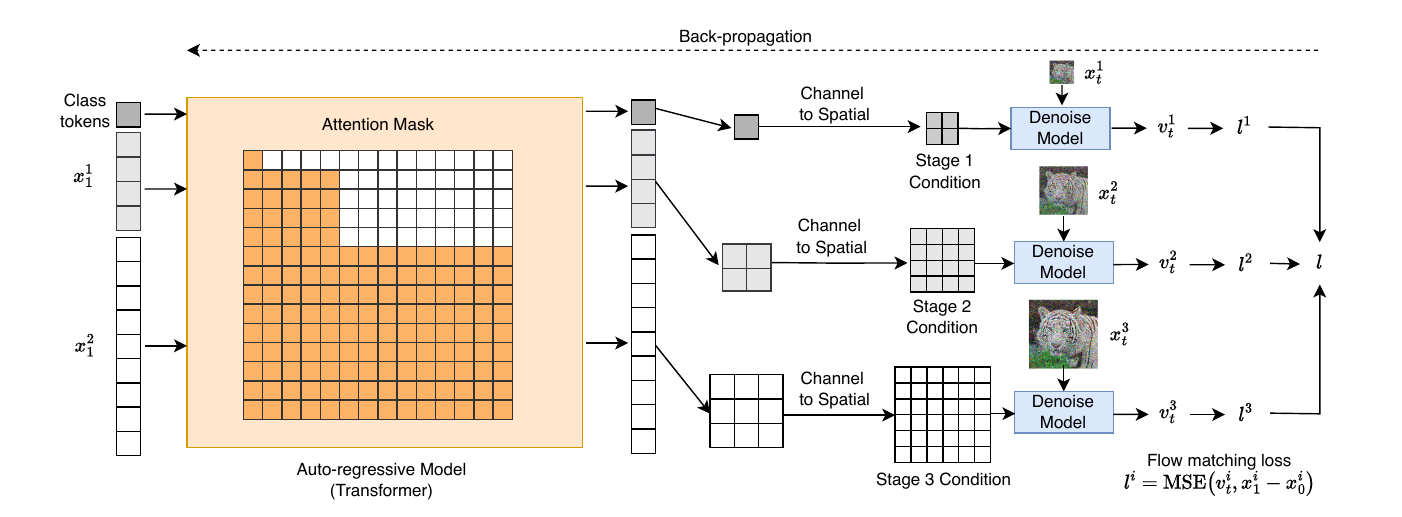}
    \caption{\textbf{Training of \ecar}. Our model combines multistage autoregressive token generation with progressive flow matching. The AR transformer (left) generates continuous token maps using a multistage causal attention mask, which are then transformed to spatial conditions for each stage. Each stage's token map conditions its corresponding flow model, enabling progressive reconstruction of image latents at different resolutions. The flow matching loss is computed between the predicted velocity and the ground truth trajectory at each stage, with back-propagation through the entire pipeline for end-to-end training.}
    \label{fig:ecar_pipeline}
    \vspace{-0.2in}
\end{figure*}

\noindent\textbf{Multi-stage Flow.}
To reduce redundant computation in early steps, we propose a multistage flow approach that operates at multiple resolutions. We interpolate the flow between the feature map and compressed low-resolution noise, progressively increasing the resolution at each stage, as shown in Fig.\ref{fig:intro}.d.

Suppose we have $S$ stages, each corresponding to a resolution level, with each stage halving the spatial dimensions of the previous one. We partition the time interval $[0,1]$ into $S$ segments:
Stage $s$ corresponds to the time interval $t \in [t_{s-1}, t_s]$, where $t_{s} = \frac{s}{S}$ for $s = 1, 2, \dots, S$.
At each stage $s$, we define the rescaled time within the stage: $\tau = \frac{t - t_{s-1}}{t_s - t_{s-1}} \in [0,1]$. 
The target feature map at the current resolution:
\begin{equation}
     \mathbf{F}^{(s)}_1 = \text{Down}\left(\mathbf{F}_1, 2^{S - s} \right),
\end{equation}
where $\mathbf{F}_1$ is the ground-truth feature map $\mathbf{F}^{(S)}$, $\text{Down}(\cdot,k)$ denote downsampling by a factor of $k$.

The initial noise feature map at the current resolution $\mathbf{F}_0^{(s)}$: 
\begin{equation} 
\mathbf{F}_0^{(s)} = 
\begin{cases} \text{Down}\left( \mathbf{F}_0, 2^{S - 1} \right), & \text{if } s = 1, \\
\text{Up}\left( \mathbf{F}_0^{(s-1)} \right), & \text{if } s > 1, \end{cases} 
\end{equation} where $\mathbf{F}_0$ is the initial noise sample (e.g., Gaussian noise), and $\text{Up}(\cdot)$ denotes upsampling by a factor of 2.
Within stage $s$, we interpolate between $\mathbf{F}_0^{(s)}$ and $\mathbf{F}_1^{(s)}$ using the rescaled time $\tau$: \begin{equation} 
\hat{\mathbf{F}}_t^{(S)} = (1 - \tau), \mathbf{F}_0^{(s)} + \tau \mathbf{F}_1^{(s)}. 
\label{eq} 
\end{equation}
This interpolation ensures that feature maps at each stage have matching dimensions, allowing us to perform computations efficiently at each resolution level.
Only the final stage ($s = S$) operates at full resolution, where no downsampling is applied: 
$\mathbf{F}_1^{(S)} = \mathbf{F}_1$.
More details about the multistage flow can be found in \citep{jin2024pyramidal} and our Appendix.

\noindent\textbf{Efficiency Benefits.} By performing computations at progressively higher resolutions, we can reduce the computational burden. Early stages operate on downsampled feature maps, which are much smaller in size, thus requiring fewer computations. Only the final stage processes the full-resolution feature map.
This multistage approach effectively reduces the overall computational cost by a factor of approximately $1/S$, assuming uniform time partitioning and resolution scaling.

\noindent\textbf{Latent Space Technique.} Even though the whole story is on how to model between continuous token map $\mathbf{M}_s$ and image feature map $\mathbf{F}^{(s)}$, the latent space technique for high-resolution image synthesis~\citep{rombach2022high} can also be easily removed out of our pipeline (\ie, directly used in the image domain $\mathbf{X}^{(s)}$ without latent space).

\subsection{Training Loss}
\label{subsec:loss}

The training of \ecar~involves optimizing both the multistage autoregressive token generation model and the multistage flow matching model. At each stage, our goal is to reconstruct the image's latent representations at varying resolutions, ensuring that the model learns to generate accurate representations progressively. Specifically, for each stage $s$:
\textbf{Continuous Token Generation}: We generate the continuous token map $\mathbf{M}_s$ using the multistage autoregressive (AR) model, as described in Sec.~\ref{subsec:h-ar}. This token map serves as the conditioning information for the flow model at stage $s$.
\textbf{Image Latent Reconstruction}: We reconstruct the image's latent representation $\mathbf{F}_s$ at resolution level $s$ using the multistage flow model. The reconstruction is conditioned on the continuous token map $\mathbf{M}_s$. Specifically, we sample $\hat{\mathbf{F}}^{(s)}$ by solving the ODE defined in Eq.~\eqref{eq:flow_sampling} using the velocity model $v_{\theta}$: 
\begin{equation} 
\hat{\mathbf{F}}^{(s)} = \text{ODE}[v_{\theta}](\mathbf{F}_0^{(s)} \mid \mathbf{M}_s), 
\end{equation} 
where $\mathbf{F}_0^{(s)}$ is the initial noise sample at stage $s$, typically a downsampled noisy image.
\textbf{Flow Matching Loss Computation}: We compute the flow matching loss $l_{\text{flow}}^{s}$ at stage $s$ using the time-dependent interpolated latent representations $\hat{\mathbf{F}}_t$ and their derivatives: 
\begin{equation} 
l_{\text{flow}}^{s} = \mathbb{E}_{(\mathbf{F}_0^{(s)}, \mathbf{F}_1^{(s)}), t \in [t_{s-1}, t_s]} \left[ \left| \frac{d \hat{\mathbf{F}}_t^{(s)}}{dt} - v_{\theta}\left( \hat{\mathbf{F}}_t^{(s)}, t \mid \mathbf{M}_s \right) \right|^2 \right], 
\end{equation} 
where $\mathbf{F}_1^{(s)}$ is the ground-truth image latent at stage $s$, and $\frac{d \hat{\mathbf{F}}_t^{(s)}}{dt} = \mathbf{F}_1^{(s)} - \mathbf{F}_0^{(s)}$ due to linear interpolation (see Fig.\ref{fig:ecar_pipeline}).

This multi-stage loss encourages the model to generate accurate representations at various resolutions, ultimately leading to high-quality final outputs.
To dynamically balance the contributions of different stages to the overall loss, we employ the GradNorm technique~\citep{chen2018gradnorm}. This adaptive weighting method automatically adjusts the stage-specific weights $w_s$ during training.

\section{Experiments}
We detail the experimental setup and model configurations in Sec.~\ref{subsec:config}, followed by an analysis of the quantitative and qualitative performance of our approach in Sec.~\ref{subsec:quant_result} and Sec.~\ref{subsec:qualit_result}. 
Finally, we demonstrate the effectiveness of each component in our model in Sec.\ref{subsec:ablation}.

\subsection{Experimental Setup}
\label{subsec:config}

We train the image generation models on ImageNet~\citep{deng2009imagenet} train dataset at 256$\times$256 resolution. 
The image tokenizer is from SDXL~\citep{podellsdxl}, which is a VAE that transform the image to the latent space with 8x reduction. 
The latent resolution is 32$\times$32 when the image resolution is 256$\times$256. 
All models are trained with the settings: AdamW optimizer with learning rate of 1e-4, no weight decay, and the batch size of 256. 
We use 8 NVIDIA A100 GPUs to train each model and the TF32 dataformat to accelerate the training process.

\begin{table}[htb]
\centering
\caption{Model configurations.}
\label{tab:model_config}
\resizebox{0.5\textwidth}{!}{
\begin{tabular}{@{}ccccc@{}}
\toprule
                           &             & E-CAR-S & E-CAR-B & E-CAR-L \\ \midrule
\multirow{3}{*}{AR}        & Layers      & 10      & 20      & 24      \\
                           & Hidden size & 768     & 1024    & 1152    \\
                           & Heads       & 6       & 16      & 16      \\ 
\bottomrule
\end{tabular}
}
\end{table}

\noindent\textbf{Model.} To demonstrate the scalability of \ecar, we trained different size of models, \ecar-\{S,B,L\}. 
The model configurations are shown in Table~\ref{tab:model_config}.
The auto-regressive model (AR) takes ViT backbone~\citep{dosovitskiy2020image}. 
There are three auto-regressive stages to generate image at latent space, and the latent resolutions of these stages are \{8,16,32\}.
For each stage, we use a diffusion model with much smaller number of Transformer layers than the auto-regressive model.
Therefore, the diffusion model is efficient to iteratively denoise on the latent space.
We use the adaptive layer normalization (adaLN)~\citep{peebles2023scalable} to incorporate the conditional information into the diffusion process.
Unlike DiT~\citep{peebles2023scalable} takes the class condition as a global information, the condition from AR model is spatially variant in E-CAR. 
That is, the condition is different at each spatial location in diffusion model.

\noindent\textbf{Metric.} We use the same
evaluation metrics as Guided Diffusion, FID~\citep{heusel2017gans}, Inception Score (IS)~\citep{salimans2016improved}, Improved Precision and Recall~\citep{kynkaanniemi2019improved}. In general, Frechet Inception Distance (FID) and Inception Score can measure the quality of the generated images, and recall score can measure the diversity of generated images.
We generate 50K images with 250 total denoising steps. Each stage takes 1/3 of the total denoising steps.



\subsection{Quantitative Results}
\label{subsec:quant_result}

To quantitatively evaluate the performance of E-CAR image generation, we trained the \ecar~{S,B,L} models for 400K iterations. 
We choose two widely used model series to compare, including DiT~\citep{peebles2023scalable} and MAR~\citep{li2024autoregressive}.


\begin{table}[tb]
\caption{Quantitative results: 256$\times$256 image generation with classifier-free guidance (CFG). IS is inception score, P is precision and R is recall.}
\centering
\label{tab:quantitative_results_cfg}
\resizebox{0.7\linewidth}{!}{
\begin{tabular}{@{}ccccccc@{}}
\toprule
Model          & Params & FLOPs & FID$\downarrow$  & IS$\uparrow$  & P$\uparrow$    & R$\uparrow$    \\ \midrule
DiT-XL/2 (7M)  & 675M   & 60T   & 2.27 & 278 & 0.83 & 0.57 \\ \midrule
MAR-B          & 208M   & 18T   & 2.31 & 282 & 0.82 & 0.57 \\
MAR-L          & 479M   & 34T   & 1.78 & 296 & 0.81 & 0.60 \\
MAR-H          & 943M   & 60T   & 1.55 & 304 & 0.81 & 0.62 \\ \midrule
E-CAR-S (400K) & 155M   & 1T    & 9.21 & 303 & 0.85 & 0.29 \\
E-CAR-B (400K) & 511M   & 3.2T  & 7.02 & 309 & 0.84 & 0.38 \\
E-CAR-L (800K) & 854M   & 5.8T  & 4.99 & 274 & 0.85 & 0.41 \\ \bottomrule
\end{tabular}
}
\vspace{-0.1in}
\end{table}

The results are shown in Table~\ref{tab:quantitative_results_cfg}.
The E-CAR-L model achieves a computational performance of 5.8 TFLOPs for generating 256$\times$256 images with classifier-free guidance (CFG), which is a mere 9.7\% of the capabilities seen in both the MAR-H and DiT-XL/2 models. E-CAR-L achieve highest inception score and precision among three types of model. Though E-CAR-L has a higher FID compared with MAR-H and DiT-XL/2, it's important to highlight that due to time constraints, our E-CAR models have only undergone 400K training iterations, significantly fewer than the 7 million iterations typical for DiT and MAR models. As shown in fig~\ref{im_right}, at 400K iterations, the model loss is continuously decreasing, suggesting that further training could potentially result in a lower FID. Our findings demonstrate that E-CAR-L significantly diminishes computational expenses while maintaining a high standard of image generation quality.

\subsection{Inference Efficiency Analysis}

\begin{table}[tb]
\caption{256$\times$256 image generation latency (s) without CFG.}
\centering
\label{tab:latency}
\resizebox{0.55\linewidth}{!}{
\begin{tabular}{@{}ccccc@{}}
\toprule
Model           & Params & 4090  & i9-13900K & Apple M2 \\ \midrule
\multicolumn{5}{c}{Batchsize=1}                         \\ \midrule
DiT-B/2  & 130M   & 1.44  & 17.94     & 8.60     \\
DiT-L/2  & 458M   & 2.73  & 59.70     & 16.50    \\
DiT-XL/2 & 675M   & 3.42  & 75.35     & 18.90    \\ \midrule
MAR-B           & 208M   & 40.61 & 146.26    & -        \\
MAR-L           & 479M   & 45.37 & 257.02    & -        \\
MAR-H           & 943M   & 59.54 & 463.62    & -        \\ \midrule
E-CAR-S  & 155M   & 0.38  & 3.92      & 2.03     \\
E-CAR-B  & 511M   & 0.68  & 10.95     & 3.35     \\
E-CAR-L  & 854M   & 0.94  & 18.22     & 4.62     \\ \midrule
\end{tabular}
}
\end{table}

To evaluate the inference efficiency of \ecar, we deploy the model on different devices including NVIDIA 4090 24G GPU, Intel i9-13900K CPU with 32G memory, and Apple M2 16G on Macbook Air. The batchsize of inference is set to 1.
The result is shown in Table~\ref{tab:latency}. The generation time of E-CAR-L model on 4090 GPU is less than 1 second. In comparison to MAR models, E-CAR models achieve a substantial speedup from 63$\times$ to 107$\times$ on NVIDIA 4090 GPUs, and 25$\times$ to 37$\times$ on Intel i9-13900K CPUs. MAR results for the Apple M2 are not listed, as the platform is not supported and encountered errors during testing. When compared with DiT models, despite having a slightly higher number of model parameters than DiT models, our models still deliver considerable acceleration on all test devices. E-CAR models achieve 3.6$\times$ - 4.0$\times$ speedup on 4090 GPUs, 4.1$\times$ - 5.5$\times$ speedup on i9-13900K CPU, and 4.1$\times$ - 4.9$\times$ speedup on Apple M2 chips.


\noindent\textbf{Efficiency of \ecar.}
\label{subsec:discuss}
Here, we provide a straightforward explanation of why \ecar~is both more efficient and effective than traditional AR image generation models.
The efficiency of \ecar~stems from three key factors: (1) Hierarchical Continuous Token Generation: As described in Sec.~\ref{subsec:h-ar}, our stage-wise approach reduces the computational complexity from $\mathcal{O}(n^3)$ to $\mathcal{O}(n^2)$, where $n$ is the number of tokens.
(2) Efficient Detokenization: The multistage flow in Sec.\ref{subsec:flow} offers a more efficient alternative to diffusion-based approaches for transforming tokens into images. Diffusion-based detokenizer needs 
(3) Parallel Processing: Unlike token-by-token generation, \ecar~can generate the entire stack of tokens for an image simultaneously. This enables parallel processing during token recovery, further enhancing efficiency.

\noindent\textbf{Effectiveness in Image Generation.}
The effectiveness of \ecar~in image generation is rooted in two principles:
(1) Continuous Representation: Images are inherently continuous signals. By operating in a continuous token space, \ecar~aligns more closely with the natural structure of images, a widely accepted inductive bias in computer vision. It echoes the finding in \citep{yu2024representation} that representation space is important for generative models' training. 
(2) Hierarchical Architecture: The stage-wise generation process of \ecar~mirrors the hierarchical nature of visual information, from coarse structures to fine details. This approach allows the model to generate images at multiple scales effectively.

\subsection{Qualitative Results}
\label{subsec:qualit_result}

\begin{figure}[!t] 
    \centering
    \includegraphics[width=0.99\linewidth]{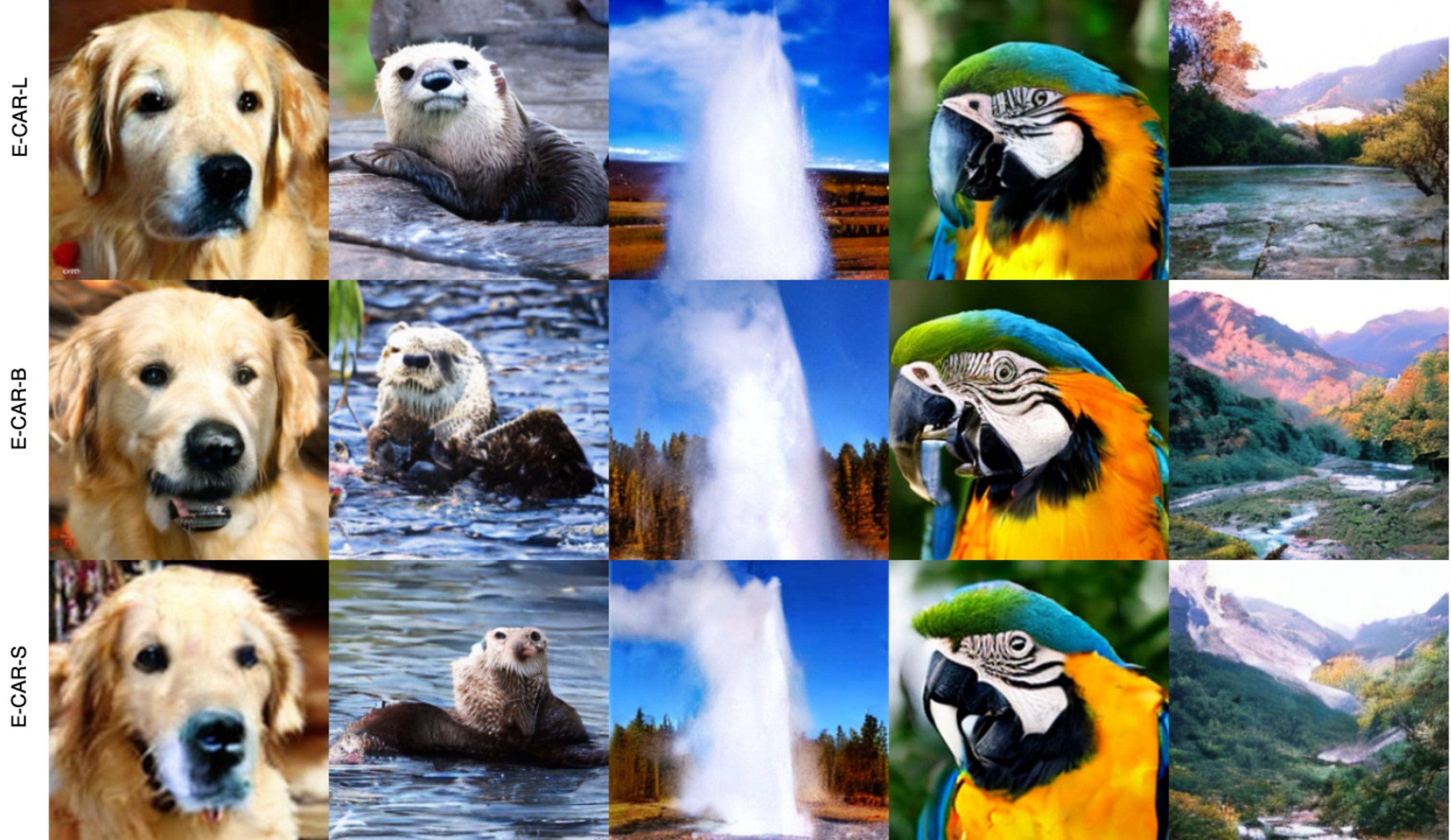}
    \caption{Samples from different models with the same noise.}
    \label{fig:samples}
\end{figure}

\begin{figure}[!tb] 
\begin{minipage}[t]{0.49\linewidth} 
    \centering 
    \includegraphics[width=\linewidth]{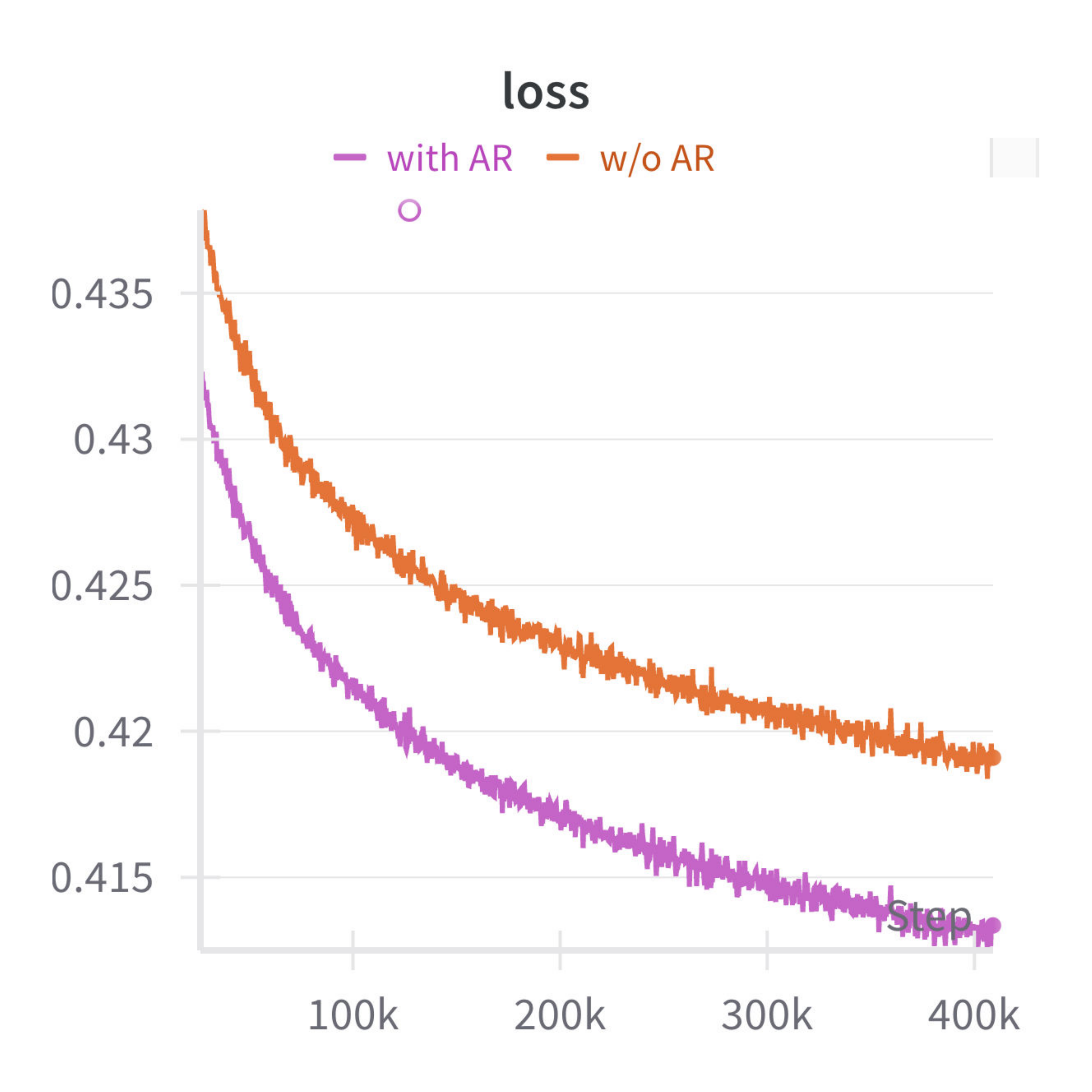} 
    \caption{Ablation study of AR.} 
    \label{fig:ablation}
    \label{im_left} 
  \end{minipage}
  \begin{minipage}[t]{0.49\linewidth} 
    \centering 
    \includegraphics[width=\linewidth]{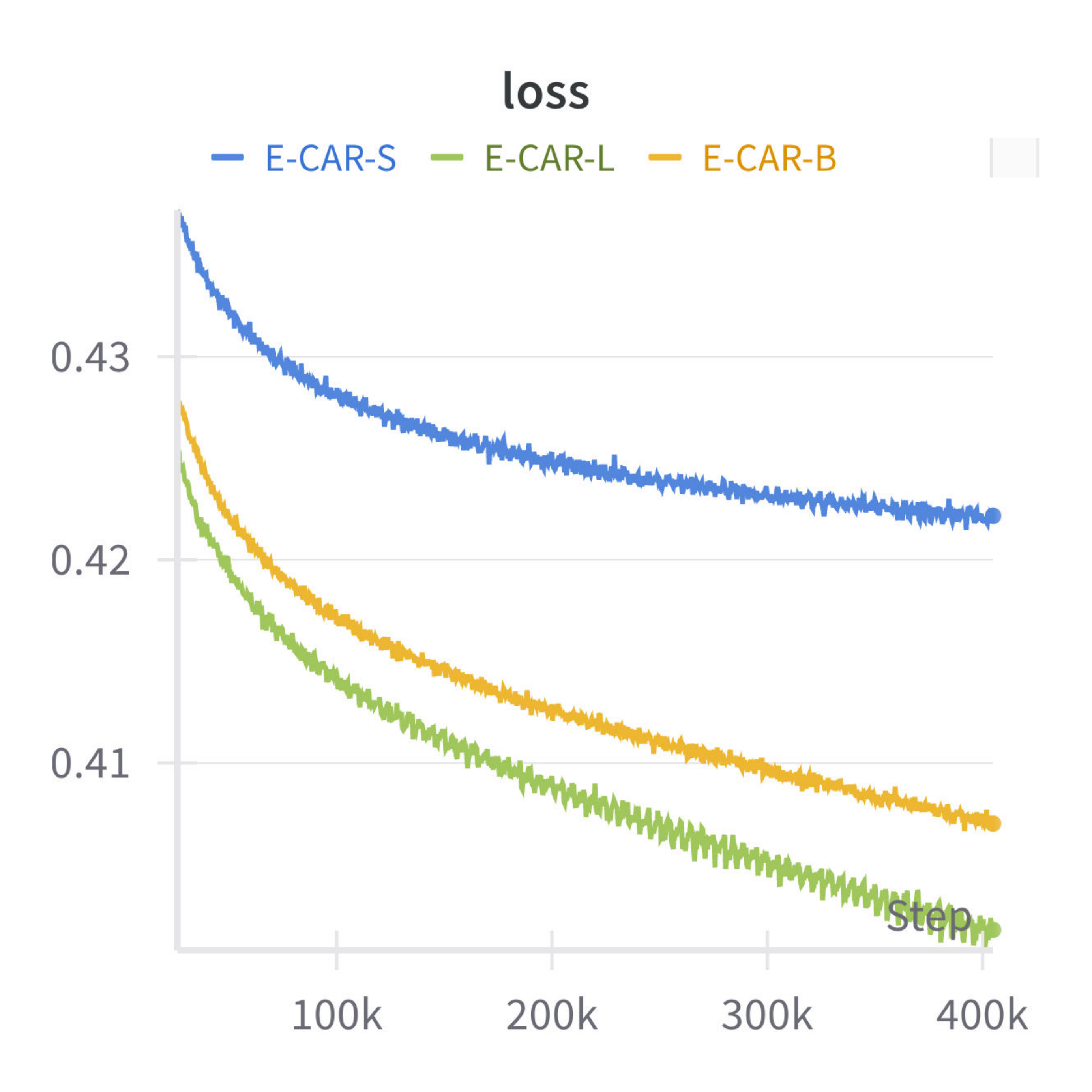} 
    \caption{Training loss.} 
    \label{im_right} 
  \end{minipage} 
\end{figure}

We show the generated images of E-CAR models with three different model size in Figure~\ref{fig:samples}. 
We can observe that our method is able to generate high quality images with 400K training rounds. Among three models, images generated by E-CAR-L has highest visual fidelity, demonstrating our method has scaling behavior.

\subsection{Ablation Study}
\label{subsec:ablation}

To show the effective of the \ecar~ architecture, we replace the layers in auto-regressive model with layers in denoising model, that is, only the low resolution image is sent to the next stage.
The results are shown in Figure~\ref{fig:ablation}.
It shows that by using the auto-regressive model, the loss can decease significantly quickly.
This indicate that the auto-regressive can plan the generation process in high-level, and the denoising model is only focused on low-level denoising process.

\section{Conclusion}

We introduce E-CAR, a method enhancing continuous AR image generation efficiency with two innovations:  1) stage-wise progressive token map generation that reduces computation by generating tokens at increasing resolutions in parallel, and 2) multistage flow-based distribution modeling that transforms only partial-denoised distributions at each stage comparing to complete denoising in normal diffusion models.
Experiments show that E-CAR achieves comparable image quality to prior work while requiring 10$\times$ fewer FLOPs and providing a 5$\times$ speedup for 256$\times$256 image generation.

\textbf{Acknowledgment}:
We would like to express our gratitude to Xuefei Ning for her invaluable guidance and support throughout this research endeavor. We are also grateful to the NICS-efc Lab and Infinigence AI for their substantial support.

\clearpage

\clearpage
{\small
\bibliographystyle{iclrbib}
\bibliography{main}

\begin{thebibliography}{51}
\providecommand{\natexlab}[1]{#1}
\providecommand{\url}[1]{\texttt{#1}}
\expandafter\ifx\csname urlstyle\endcsname\relax
  \providecommand{\doi}[1]{doi: #1}\else
  \providecommand{\doi}{doi: \begingroup \urlstyle{rm}\Url}\fi

\bibitem[Achiam et~al.(2023)Achiam, Adler, Agarwal, Ahmad, Akkaya, Aleman, Almeida, Altenschmidt, Altman, Anadkat, et~al.]{achiam2023gpt}
Josh Achiam, Steven Adler, Sandhini Agarwal, Lama Ahmad, Ilge Akkaya, Florencia~Leoni Aleman, Diogo Almeida, Janko Altenschmidt, Sam Altman, Shyamal Anadkat, et~al.
\newblock Gpt-4 technical report.
\newblock \emph{arXiv preprint arXiv:2303.08774}, 2023.

\bibitem[Chen et~al.(2020)Chen, Radford, Child, Wu, Jun, Luan, and Sutskever]{chen2020generative}
Mark Chen, Alec Radford, Rewon Child, Jeffrey Wu, Heewoo Jun, David Luan, and Ilya Sutskever.
\newblock Generative pretraining from pixels.
\newblock In \emph{ICML}, 2020.

\bibitem[Chen et~al.(2018{\natexlab{a}})Chen, Mishra, Rohaninejad, and Abbeel]{chen2018pixelsnail}
Xi~Chen, Nikhil Mishra, Mostafa Rohaninejad, and Pieter Abbeel.
\newblock Pixelsnail: An improved autoregressive generative model.
\newblock In \emph{ICML}, 2018{\natexlab{a}}.

\bibitem[Chen et~al.(2018{\natexlab{b}})Chen, Badrinarayanan, Lee, and Rabinovich]{chen2018gradnorm}
Zhao Chen, Vijay Badrinarayanan, Chen-Yu Lee, and Andrew Rabinovich.
\newblock Gradnorm: Gradient normalization for adaptive loss balancing in deep multitask networks.
\newblock In \emph{ICML}, 2018{\natexlab{b}}.

\bibitem[Croitoru et~al.(2023)Croitoru, Hondru, Ionescu, and Shah]{croitoru2023diffusion}
Florinel-Alin Croitoru, Vlad Hondru, Radu~Tudor Ionescu, and Mubarak Shah.
\newblock Diffusion models in vision: A survey.
\newblock \emph{TPAMI}, 2023.

\bibitem[Deng et~al.(2009)Deng, Dong, Socher, Li, Li, and Fei-Fei]{deng2009imagenet}
Jia Deng, Wei Dong, Richard Socher, Li-Jia Li, Kai Li, and Li~Fei-Fei.
\newblock Imagenet: A large-scale hierarchical image database.
\newblock In \emph{CVPR}, 2009.

\bibitem[Dosovitskiy(2020)]{dosovitskiy2020image}
Alexey Dosovitskiy.
\newblock An image is worth 16x16 words: Transformers for image recognition at scale.
\newblock \emph{arXiv preprint arXiv:2010.11929}, 2020.

\bibitem[Esser et~al.(2021)Esser, Rombach, and Ommer]{esser2021taming}
Patrick Esser, Robin Rombach, and Bjorn Ommer.
\newblock Taming transformers for high-resolution image synthesis.
\newblock In \emph{CVPR}, 2021.

\bibitem[Esser et~al.(2024)Esser, Kulal, Blattmann, Entezari, M{\"u}ller, Saini, Levi, Lorenz, Sauer, Boesel, et~al.]{esser2024scaling}
Patrick Esser, Sumith Kulal, Andreas Blattmann, Rahim Entezari, Jonas M{\"u}ller, Harry Saini, Yam Levi, Dominik Lorenz, Axel Sauer, Frederic Boesel, et~al.
\newblock Scaling rectified flow transformers for high-resolution image synthesis.
\newblock In \emph{ICML}, 2024.

\bibitem[Fan et~al.(2024)Fan, Li, Qin, Li, Sun, Rubinstein, Sun, He, and Tian]{fan2024fluid}
Lijie Fan, Tianhong Li, Siyang Qin, Yuanzhen Li, Chen Sun, Michael Rubinstein, Deqing Sun, Kaiming He, and Yonglong Tian.
\newblock Fluid: Scaling autoregressive text-to-image generative models with continuous tokens.
\newblock \emph{arXiv preprint arXiv:2410.13863}, 2024.

\bibitem[Gregor et~al.(2014)Gregor, Danihelka, Mnih, Blundell, and Wierstra]{gregor2014deep}
Karol Gregor, Ivo Danihelka, Andriy Mnih, Charles Blundell, and Daan Wierstra.
\newblock Deep autoregressive networks.
\newblock In \emph{ICML}, 2014.

\bibitem[Heusel et~al.(2017)Heusel, Ramsauer, Unterthiner, Nessler, and Hochreiter]{heusel2017gans}
Martin Heusel, Hubert Ramsauer, Thomas Unterthiner, Bernhard Nessler, and Sepp Hochreiter.
\newblock Gans trained by a two time-scale update rule converge to a local nash equilibrium.
\newblock \emph{NeurIPS}, 2017.

\bibitem[Ho et~al.(2020)Ho, Jain, and Abbeel]{ho2020denoising}
Jonathan Ho, Ajay Jain, and Pieter Abbeel.
\newblock Denoising diffusion probabilistic models.
\newblock \emph{NeurIPS}, 2020.

\bibitem[Hoffmann et~al.(2022)Hoffmann, Borgeaud, Mensch, Buchatskaya, Cai, Rutherford, Casas, Hendricks, Welbl, Clark, et~al.]{hoffmann2022training}
Jordan Hoffmann, Sebastian Borgeaud, Arthur Mensch, Elena Buchatskaya, Trevor Cai, Eliza Rutherford, Diego de~Las Casas, Lisa~Anne Hendricks, Johannes Welbl, Aidan Clark, et~al.
\newblock Training compute-optimal large language models.
\newblock \emph{arXiv preprint arXiv:2203.15556}, 2022.

\bibitem[Jiang et~al.(2023)Jiang, Sablayrolles, Mensch, Bamford, Chaplot, Casas, Bressand, Lengyel, Lample, Saulnier, et~al.]{jiang2023mistral}
Albert~Q Jiang, Alexandre Sablayrolles, Arthur Mensch, Chris Bamford, Devendra~Singh Chaplot, Diego de~las Casas, Florian Bressand, Gianna Lengyel, Guillaume Lample, Lucile Saulnier, et~al.
\newblock Mistral 7b.
\newblock \emph{arXiv preprint arXiv:2310.06825}, 2023.

\bibitem[Jin et~al.(2024)Jin, Sun, Li, Xu, Jiang, Zhuang, Huang, Song, Mu, and Lin]{jin2024pyramidal}
Yang Jin, Zhicheng Sun, Ningyuan Li, Kun Xu, Hao Jiang, Nan Zhuang, Quzhe Huang, Yang Song, Yadong Mu, and Zhouchen Lin.
\newblock Pyramidal flow matching for efficient video generative modeling.
\newblock \emph{arXiv preprint arXiv:2410.05954}, 2024.

\bibitem[Kaplan et~al.(2020)Kaplan, McCandlish, Henighan, Brown, Chess, Child, Gray, Radford, Wu, and Amodei]{kaplan2020scaling}
Jared Kaplan, Sam McCandlish, Tom Henighan, Tom~B Brown, Benjamin Chess, Rewon Child, Scott Gray, Alec Radford, Jeffrey Wu, and Dario Amodei.
\newblock Scaling laws for neural language models.
\newblock \emph{arXiv preprint arXiv:2001.08361}, 2020.

\bibitem[Kynk{\"a}{\"a}nniemi et~al.(2019)Kynk{\"a}{\"a}nniemi, Karras, Laine, Lehtinen, and Aila]{kynkaanniemi2019improved}
Tuomas Kynk{\"a}{\"a}nniemi, Tero Karras, Samuli Laine, Jaakko Lehtinen, and Timo Aila.
\newblock Improved precision and recall metric for assessing generative models.
\newblock In \emph{NeurIPS}, 2019.

\bibitem[Lee et~al.(2022)Lee, Kim, Kim, Cho, and Han]{lee2022autoregressive}
Doyup Lee, Chiheon Kim, Saehoon Kim, Minsu Cho, and Wook-Shin Han.
\newblock Autoregressive image generation using residual quantization.
\newblock In \emph{Proceedings of the IEEE/CVF Conference on Computer Vision and Pattern Recognition}, pp.\  11523--11532, 2022.

\bibitem[Li et~al.(2024)Li, Tian, Li, Deng, and He]{li2024autoregressive}
Tianhong Li, Yonglong Tian, He~Li, Mingyang Deng, and Kaiming He.
\newblock Autoregressive image generation without vector quantization.
\newblock \emph{arXiv preprint arXiv:2406.11838}, 2024.

\bibitem[Lipman et~al.(2022)Lipman, Chen, Ben-Hamu, Nickel, and Le]{lipman2022flow}
Yaron Lipman, Ricky~TQ Chen, Heli Ben-Hamu, Maximilian Nickel, and Matt Le.
\newblock Flow matching for generative modeling.
\newblock \emph{arXiv preprint arXiv:2210.02747}, 2022.

\bibitem[Liu et~al.(2024{\natexlab{a}})Liu, Zhao, Zhuo, Lin, Qiao, Li, and Gao]{liu2024lumina}
Dongyang Liu, Shitian Zhao, Le~Zhuo, Weifeng Lin, Yu~Qiao, Hongsheng Li, and Peng Gao.
\newblock Lumina-mgpt: Illuminate flexible photorealistic text-to-image generation with multimodal generative pretraining.
\newblock \emph{arXiv preprint arXiv:2408.02657}, 2024{\natexlab{a}}.

\bibitem[Liu(2022)]{liu2022rectified}
Qiang Liu.
\newblock Rectified flow: A marginal preserving approach to optimal transport.
\newblock \emph{arXiv preprint arXiv:2209.14577}, 2022.

\bibitem[Liu et~al.(2022)Liu, Gong, and Liu]{liu2022flow}
Xingchao Liu, Chengyue Gong, and Qiang Liu.
\newblock Flow straight and fast: Learning to generate and transfer data with rectified flow.
\newblock \emph{arXiv preprint arXiv:2209.03003}, 2022.

\bibitem[Liu et~al.(2024{\natexlab{b}})Liu, Zhang, Ma, Peng, et~al.]{liu2023instaflow}
Xingchao Liu, Xiwen Zhang, Jianzhu Ma, Jian Peng, et~al.
\newblock Instaflow: One step is enough for high-quality diffusion-based text-to-image generation.
\newblock In \emph{ICLR}, 2024{\natexlab{b}}.

\bibitem[Ma et~al.(2024)Ma, Goldstein, Albergo, Boffi, Vanden-Eijnden, and Xie]{ma2024sit}
Nanye Ma, Mark Goldstein, Michael~S Albergo, Nicholas~M Boffi, Eric Vanden-Eijnden, and Saining Xie.
\newblock Sit: Exploring flow and diffusion-based generative models with scalable interpolant transformers.
\newblock \emph{arXiv preprint arXiv:2401.08740}, 2024.

\bibitem[Peebles \& Xie(2023)Peebles and Xie]{peebles2023scalable}
William Peebles and Saining Xie.
\newblock Scalable diffusion models with transformers.
\newblock In \emph{Proceedings of the IEEE/CVF International Conference on Computer Vision}, pp.\  4195--4205, 2023.

\bibitem[Pernias et~al.(2023)Pernias, Rampas, Richter, Pal, and Aubreville]{pernias2023wurstchen}
Pablo Pernias, Dominic Rampas, Mats~L Richter, Christopher~J Pal, and Marc Aubreville.
\newblock W{\"u}rstchen: An efficient architecture for large-scale text-to-image diffusion models.
\newblock \emph{arXiv preprint arXiv:2306.00637}, 2023.

\bibitem[Podell et~al.()Podell, English, Lacey, Blattmann, Dockhorn, M{\"u}ller, Penna, and Rombach]{podellsdxl}
Dustin Podell, Zion English, Kyle Lacey, Andreas Blattmann, Tim Dockhorn, Jonas M{\"u}ller, Joe Penna, and Robin Rombach.
\newblock Sdxl: Improving latent diffusion models for high-resolution image synthesis.
\newblock In \emph{The Twelfth International Conference on Learning Representations}.

\bibitem[Radford(2018)]{radford2018improving}
Alec Radford.
\newblock Improving language understanding by generative pre-training.
\newblock 2018.

\bibitem[Razavi et~al.(2019)Razavi, Van~den Oord, and Vinyals]{razavi2019generating}
Ali Razavi, Aaron Van~den Oord, and Oriol Vinyals.
\newblock Generating diverse high-fidelity images with vq-vae-2.
\newblock \emph{Advances in neural information processing systems}, 32, 2019.

\bibitem[Rombach et~al.(2022)Rombach, Blattmann, Lorenz, Esser, and Ommer]{rombach2022high}
Robin Rombach, Andreas Blattmann, Dominik Lorenz, Patrick Esser, and Bj{\"o}rn Ommer.
\newblock High-resolution image synthesis with latent diffusion models.
\newblock In \emph{CVPR}, 2022.

\bibitem[Saharia et~al.(2022)Saharia, Chan, Saxena, Li, Whang, Denton, Ghasemipour, Gontijo~Lopes, Karagol~Ayan, Salimans, et~al.]{saharia2022photorealistic}
Chitwan Saharia, William Chan, Saurabh Saxena, Lala Li, Jay Whang, Emily~L Denton, Kamyar Ghasemipour, Raphael Gontijo~Lopes, Burcu Karagol~Ayan, Tim Salimans, et~al.
\newblock Photorealistic text-to-image diffusion models with deep language understanding.
\newblock \emph{NeurIPS}, 2022.

\bibitem[Salimans et~al.(2016)Salimans, Goodfellow, Zaremba, Cheung, Radford, and Chen]{salimans2016improved}
Tim Salimans, Ian Goodfellow, Wojciech Zaremba, Vicki Cheung, Alec Radford, and Xi~Chen.
\newblock Improved techniques for training gans.
\newblock \emph{Advances in neural information processing systems}, 29, 2016.

\bibitem[Sohl-Dickstein et~al.(2015)Sohl-Dickstein, Weiss, Maheswaranathan, and Ganguli]{sohl2015deep}
Jascha Sohl-Dickstein, Eric Weiss, Niru Maheswaranathan, and Surya Ganguli.
\newblock Deep unsupervised learning using nonequilibrium thermodynamics.
\newblock In \emph{ICML}, 2015.

\bibitem[Song et~al.(2020{\natexlab{a}})Song, Meng, and Ermon]{song2020denoising}
Jiaming Song, Chenlin Meng, and Stefano Ermon.
\newblock Denoising diffusion implicit models.
\newblock \emph{arXiv preprint arXiv:2010.02502}, 2020{\natexlab{a}}.

\bibitem[Song et~al.(2020{\natexlab{b}})Song, Sohl-Dickstein, Kingma, Kumar, Ermon, and Poole]{song2020score}
Yang Song, Jascha Sohl-Dickstein, Diederik~P Kingma, Abhishek Kumar, Stefano Ermon, and Ben Poole.
\newblock Score-based generative modeling through stochastic differential equations.
\newblock \emph{ICLR}, 2020{\natexlab{b}}.

\bibitem[Sun et~al.(2024)Sun, Jiang, Chen, Zhang, Peng, Luo, and Yuan]{sun2024autoregressive}
Peize Sun, Yi~Jiang, Shoufa Chen, Shilong Zhang, Bingyue Peng, Ping Luo, and Zehuan Yuan.
\newblock Autoregressive model beats diffusion: Llama for scalable image generation.
\newblock \emph{arXiv preprint arXiv:2406.06525}, 2024.

\bibitem[Team(2024)]{team2024chameleon}
Chameleon Team.
\newblock Chameleon: Mixed-modal early-fusion foundation models.
\newblock \emph{arXiv preprint arXiv:2405.09818}, 2024.

\bibitem[Tian et~al.(2024)Tian, Jiang, Yuan, Peng, and Wang]{tian2024visual}
Keyu Tian, Yi~Jiang, Zehuan Yuan, Bingyue Peng, and Liwei Wang.
\newblock Visual autoregressive modeling: Scalable image generation via next-scale prediction.
\newblock \emph{arXiv preprint arXiv:2404.02905}, 2024.

\bibitem[Touvron et~al.(2023{\natexlab{a}})Touvron, Lavril, Izacard, Martinet, Lachaux, Lacroix, Rozi{\`e}re, Goyal, Hambro, Azhar, et~al.]{touvron2023llama}
Hugo Touvron, Thibaut Lavril, Gautier Izacard, Xavier Martinet, Marie-Anne Lachaux, Timoth{\'e}e Lacroix, Baptiste Rozi{\`e}re, Naman Goyal, Eric Hambro, Faisal Azhar, et~al.
\newblock Llama: Open and efficient foundation language models.
\newblock \emph{arXiv preprint arXiv:2302.13971}, 2023{\natexlab{a}}.

\bibitem[Touvron et~al.(2023{\natexlab{b}})Touvron, Martin, Stone, Albert, Almahairi, Babaei, Bashlykov, Batra, Bhargava, Bhosale, et~al.]{touvron2023llama2}
Hugo Touvron, Louis Martin, Kevin Stone, Peter Albert, Amjad Almahairi, Yasmine Babaei, Nikolay Bashlykov, Soumya Batra, Prajjwal Bhargava, Shruti Bhosale, et~al.
\newblock Llama 2: Open foundation and fine-tuned chat models.
\newblock \emph{arXiv preprint arXiv:2307.09288}, 2023{\natexlab{b}}.

\bibitem[Tschannen et~al.(2023)Tschannen, Eastwood, and Mentzer]{tschannen2023givt}
Michael Tschannen, Cian Eastwood, and Fabian Mentzer.
\newblock Givt: Generative infinite-vocabulary transformers.
\newblock \emph{arXiv preprint arXiv:2312.02116}, 2023.

\bibitem[Van Den~Oord et~al.(2016)Van Den~Oord, Kalchbrenner, and Kavukcuoglu]{van2016pixel}
A{\"a}ron Van Den~Oord, Nal Kalchbrenner, and Koray Kavukcuoglu.
\newblock Pixel recurrent neural networks.
\newblock In \emph{ICML}, 2016.

\bibitem[Vaswani(2017)]{vaswani2017attention}
A~Vaswani.
\newblock Attention is all you need.
\newblock \emph{NeurIPS}, 2017.

\bibitem[Xie et~al.(2024)Xie, Zu, Zhao, Su, Liu, Shi, Li, Zhang, and Ma]{xie2024towards}
Shenghao Xie, Wenqiang Zu, Mingyang Zhao, Duo Su, Shilong Liu, Ruohua Shi, Guoqi Li, Shanghang Zhang, and Lei Ma.
\newblock Towards unifying understanding and generation in the era of vision foundation models: A survey from the autoregression perspective.
\newblock \emph{arXiv preprint arXiv:2410.22217}, 2024.

\bibitem[Yu et~al.(2023{\natexlab{a}})Yu, Cheng, Sohn, Lezama, Zhang, Chang, Hauptmann, Yang, Hao, Essa, et~al.]{yu2023magvit}
Lijun Yu, Yong Cheng, Kihyuk Sohn, Jos{\'e} Lezama, Han Zhang, Huiwen Chang, Alexander~G Hauptmann, Ming-Hsuan Yang, Yuan Hao, Irfan Essa, et~al.
\newblock Magvit: Masked generative video transformer.
\newblock In \emph{CVPR}, 2023{\natexlab{a}}.

\bibitem[Yu et~al.(2023{\natexlab{b}})Yu, Lezama, Gundavarapu, Versari, Sohn, Minnen, Cheng, Gupta, Gu, Hauptmann, et~al.]{yu2023language}
Lijun Yu, Jos{\'e} Lezama, Nitesh~B Gundavarapu, Luca Versari, Kihyuk Sohn, David Minnen, Yong Cheng, Agrim Gupta, Xiuye Gu, Alexander~G Hauptmann, et~al.
\newblock Language model beats diffusion--tokenizer is key to visual generation.
\newblock \emph{arXiv preprint arXiv:2310.05737}, 2023{\natexlab{b}}.

\bibitem[Yu et~al.(2024)Yu, Kwak, Jang, Jeong, Huang, Shin, and Xie]{yu2024representation}
Sihyun Yu, Sangkyung Kwak, Huiwon Jang, Jongheon Jeong, Jonathan Huang, Jinwoo Shin, and Saining Xie.
\newblock Representation alignment for generation: Training diffusion transformers is easier than you think.
\newblock \emph{arXiv preprint arXiv:2410.06940}, 2024.

\bibitem[Yuan et~al.(2024)Yuan, Shang, Zhou, Dong, Xue, Wu, Li, Gu, Lee, Yan, et~al.]{yuan2024llm}
Zhihang Yuan, Yuzhang Shang, Yang Zhou, Zhen Dong, Chenhao Xue, Bingzhe Wu, Zhikai Li, Qingyi Gu, Yong~Jae Lee, Yan Yan, et~al.
\newblock Llm inference unveiled: Survey and roofline model insights.
\newblock \emph{arXiv preprint arXiv:2402.16363}, 2024.

\bibitem[Zhou et~al.(2024)Zhou, Yu, Babu, Tirumala, Yasunaga, Shamis, Kahn, Ma, Zettlemoyer, and Levy]{zhou2024transfusion}
Chunting Zhou, Lili Yu, Arun Babu, Kushal Tirumala, Michihiro Yasunaga, Leonid Shamis, Jacob Kahn, Xuezhe Ma, Luke Zettlemoyer, and Omer Levy.
\newblock Transfusion: Predict the next token and diffuse images with one multi-modal model.
\newblock \emph{arXiv preprint arXiv:2408.11039}, 2024.

\end{thebibliography}
}

\clearpage
\appendix
\section{Appendix}

\subsection{Training More Iterations}

We have extended the training of E-CAR-L to 800,000 iterations, resulting in a Fréchet Inception Distance (FID) of 4.9. This represents a significant improvement compared to the 400,000 iterations, which achieved an FID of 6.5. However, we have observed that the loss has not yet converged and continues to decrease steadily. Therefore, we are continuing the training of this model for more iterations, as we believe the results will improve further.

\subsection{Token Generation Efficiency Analysis}

For a standard self-attention transformer, the time complexity of AR generation is $O(n^3)$, where $n$ is the total number of image tokens.

In AR generation, tokens are generated one at a time. For the $i$-th token ($1 \leq i \leq n$), the model computes attention scores between the new token and all previously generated $i-1$ tokens. The attention computation scales quadratically with the sequence length, so the time complexity for generating the $i$-th token is $O(i^2)$.

The total time complexity is the sum over all $n$ tokens:

\begin{align*}
T_{\text{AR}} &= \sum_{i=1}^{n} O(i^2) \\
&= O\left( \sum_{i=1}^{n} i^2 \right) \\
&= O\left( \frac{n(n+1)(2n+1)}{6} \right) \\
&= O(n^3).
\end{align*}

For a standard self-attention transformer using the multistage method with a resolution schedule, the time complexity is $O(n^2)$.

In Multistage AR generation, tokens are generated in stages with increasing resolutions. Define a sequence of stages $(n_1, n_2, \ldots, n_K)$, where $n_k$ is the number of tokens at stage $k$, and $K$ is the total number of stages.

Assume that the number of tokens doubles at each stage (for simplicity), starting from $n_1 = 1$:

\[
n_k = 2^{k-1}, \quad \text{for } 1 \leq k \leq K.
\]

The cumulative number of tokens up to stage $k$ is:

\[
S_k = \sum_{i=1}^{k} n_i = 2^{k} - 1.
\]

Since the total number of tokens is $n$, we have:

\[
n = S_K = 2^{K} - 1 \implies K = \log_2(n + 1).
\]

At each stage $k$, the time complexity is proportional to $S_k^2$ (due to the quadratic scaling of attention computation):

\[
T_{\text{MultiAR}} = \sum_{k=1}^{K} O(S_k^2).
\]

Substitute $S_k = 2^{k} - 1$:

\begin{align*}
T_{\text{MultiAR}} &= O\left( \sum_{k=1}^{K} (2^{k} - 1)^2 \right) \\
&= O\left( \sum_{k=1}^{K} \left( 2^{2k} - 2^{k+1} + 1 \right) \right) \\
&= O\left( \sum_{k=1}^{K} 4^{k} - 2^{k+1} + 1 \right).
\end{align*}

Compute each term separately and combine the sums:

\begin{align*}
T_{\text{MultiAR}} &= O\left( \frac{4^{K+1} - 4}{3} - 2^{K+2} + 4 + K \right) \\
&= O\left( \frac{4^{K+1}}{3} - 2^{K+2} + K + \text{constants} \right).
\end{align*}

Since $K = \log_2(n + 1)$, we have:

\[
4^{K+1} = 2^{2(K+1)} = 2^{2(\log_2(n + 1) + 1)} = 4(n + 1)^2.
\]

Similarly:

\[
2^{K+2} = 2^{\log_2(n + 1) + 2} = 4(n + 1).
\]

Therefore, the dominant terms in $T_{\text{MultiAR}}$ are:

\begin{align*}
T_{\text{MultiAR}} &= O\left( \frac{4(n + 1)^2}{3} - 4(n + 1) + \log_2(n + 1) \right) \\
&= O\left( n^2 \right).
\end{align*}

Comparing the time complexities:

\[
\frac{T_{\text{AR}}}{T_{\text{MultiAR}}} = \frac{O(n^3)}{O(n^2)} = O(n).
\]

Thus, the multistage AR method is faster than the AR method by a factor of $O(n)$ when generating an image with $n$ tokens. The modeling and proof are inspired by VAR~\citep{tian2024visual}.

\subsection{Upsample-Renoise in Multi-stage Flow}

In the inference phase, because the interpolation of $\hat{\mathbf{F}}_t$ involves feature maps of varying dimensions across different stages, we employ an Upsampling and Re-noising module inspired by Pyramidal Flow Matching~\citep{jin2024pyramidal}. This module ensures that the Gaussian distributions are matched at each transition point by applying a linear transformation to the upsampled results. Specifically, when moving from a lower-resolution stage to a higher-resolution one, we upsample the feature map and add Gaussian noise to match the statistical properties required for flow matching. This process facilitates smooth transitions between stages and maintains consistency in the multistage flow framework.

\end{document}